\documentclass[runningheads]{llncs}

\usepackage[mobile]{eccv}

\usepackage{eccvabbrv}

\usepackage{graphicx}
\usepackage{booktabs}
\usepackage[inline]{enumitem}
\usepackage[numbers, sort, comma, square]{natbib}
\usepackage{subfiles}
\usepackage{graphicx}
\usepackage{amsmath}
\usepackage{wrapfig}
\usepackage{caption}
\usepackage{subcaption}
\usepackage{multirow}
\usepackage{bbding}
\usepackage{arydshln}
\usepackage{relsize}
\usepackage{amssymb, bm}
\usepackage{wrapfig}
\usepackage{tikz}
\usepackage{todonotes}
\usepackage[normalem]{ulem}
\DeclareMathOperator*{\argmax}{arg\,max}
\usepackage{algorithm}
\usepackage{listings}
\usepackage[accsupp]{axessibility}  

\usepackage{hyperref}

\usepackage{orcidlink}
\begin{document}
\newtheorem{defn}{Definition}[section]
\title{Simple Unsupervised Knowledge Distillation With Space Similarity} 

\titlerunning{Simple UKD With Space Similarity}

\author{Aditya Singh\inst{1}\orcidlink{0009-0005-0452-8579} \and
Haohan Wang\inst{2}}

\authorrunning{Singh \& Wang}

\institute{
\email{aditya.91.singh@gmail.com} \\
\and 
School of Information Sciences, University of Illinois Urbana-Champaign, Champaign, IL, USA \\
\email{haohanw@illinois.edu}}

\maketitle

\begin{abstract}
As per recent studies, Self-supervised learning (SSL) does not readily extend to smaller architectures. One direction to mitigate this shortcoming while simultaneously training a smaller network without labels is to adopt unsupervised knowledge distillation (UKD). Existing UKD approaches handcraft preservation worthy inter/intra sample relationships between the teacher and its student. However, this may overlook/ignore other key relationships present in the mapping of a teacher. In this paper, instead of heuristically constructing preservation worthy relationships between samples, we directly motivate the student to model the teacher's embedding manifold. If the mapped manifold is similar, all inter/intra sample relationships are indirectly conserved. We first demonstrate that prior methods cannot preserve teacher's latent manifold due to their sole reliance on $L_2$ normalised embedding features. Subsequently, we propose a simple objective to capture the lost information due to normalisation. Our proposed loss component, termed \textbf{space similarity}, motivates each dimension of a student's feature space to be similar to the corresponding dimension of its teacher. We perform extensive experiments demonstrating strong performance of our proposed approach on various benchmarks. 
\end{abstract}

\section{Introduction}
\label{sect:introduction}
In recent years, the development of self-supervised learning (SSL) has allowed networks to be trained on larger datasets without labels, leading to generic representations that are task agnostic and achieve superior downstream performances once fine-tuned \citep{ssl_0, ssl_1, ssl_2}. 
As a result, SSL is an active area of reasearch. For real-time inference, such as in the domain of autonomous driving, industrial automation \etc often small sized networks are deployed.
However, these networks do not readily benefit from SSL due to their smaller number of parameters, which can hinder their ability to learn underlying discriminative representations effectively \cite{kd_seed}.  

To address this issue, \citet{kd_seed} propose an unsupervised knowledge distillation (UKD) framework called SEED that allows smaller networks to take advantage of the large amount of data for pre-training. Since the introduction of SEED, many other approaches have followed suite \cite{kd_bingo, kd_compress}. A common theme amongst many existing UKD methods is that they rely on manually constructing and conserving similarity relationships between training samples. Often this preservation is achieved via.\ leveraging an embedding queue ($\sim\!100k$ in length) which stores features of previously encountered training samples. This conservation of relationships (either at a local or global scale) can be perceived as an attempt to mimic the teacher's manifold. A perfect replication of a teacher's embedding manifold will imply that the pairwise similarity relationships are preserved by the student. Moreover, modelling of the manifold will also preserve those relationships/properties which were otherwise overlooked. However, we observe that, due to the normalization of embedding features, an essential step in existing training recipes, the modelling of teacher's manifold by existing approaches is imperfect (details in \cref{sect:motivation}).

\begin{figure*}[t]
    \centering
    \captionsetup{font=small}
    \footnotesize
    \begin{subfigure}{1\linewidth}
    \centering
    \includegraphics[width=0.9\textwidth]{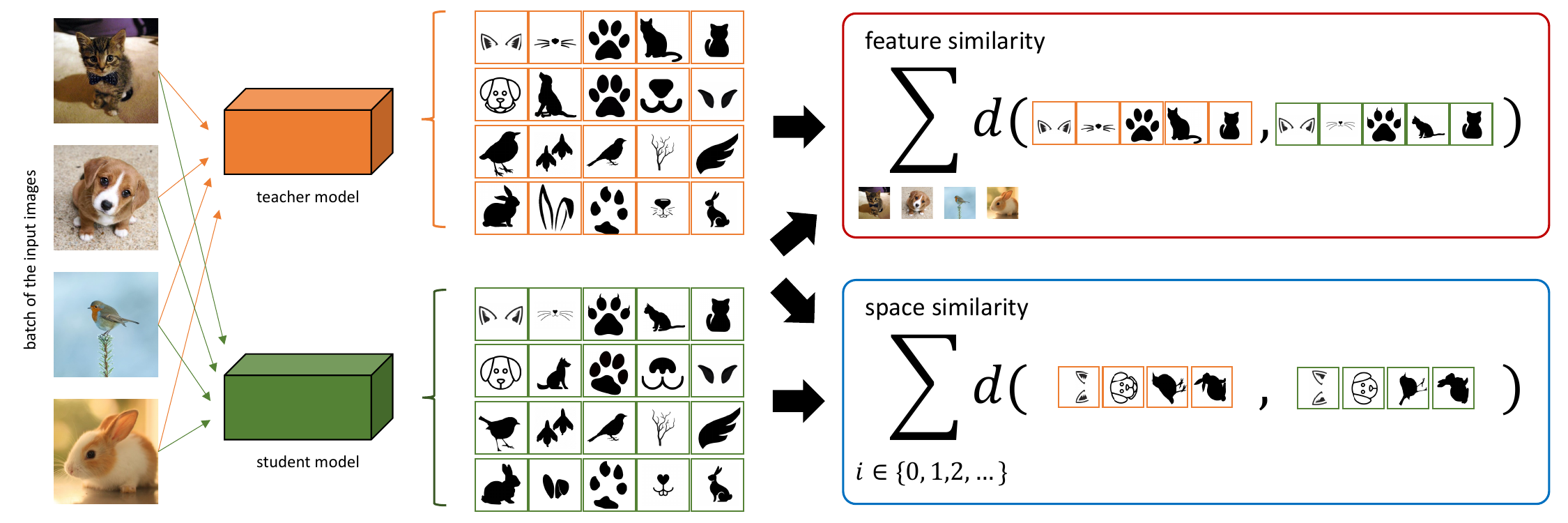}
    \end{subfigure}
    \caption{\small{\textbf{The proposed CoSS (feature similarity + space similarity) distillation framework.} In the graphic, we demonstrate similarity of one pair of corresponding feature dimension being maximised. We perform this maximisation for every corresponding pair for the teacher and student.} 
    }
    \label{fig:coss}
\end{figure*}

In this paper, we hypothesize that the knowledge of the teacher is not only in the relationship of the samples but also the manner in which these are mapped onto the latent space \ie the embedding manifold. An alignment of the embedded manifold will imply that the student has learnt to map inputs in the `same' way as the teacher onto the latent space which will indirectly preserve relationships of interest.

We find that due to sole reliance on $L_2$ normalized embedding features, existing methods cannot model teacher's embedding manifold. This is because normalization is a non-invertible mapping which eliminates the information and structure held by the original manifold. As a solution, in this paper, we propose a simple \textbf{space similarity} objective which works in conjunction with a traditional cosine similarity loss computed on the features. In space similarity, for each feature dimension of a student, we aim to maximise its similarity to the corresponding feature dimension of the teacher. Therefore, space similarity preserves the spatial information, while, the conventional cosine similarity of features ensures that the representations learnt are consistent and aligned. \cref{fig:coss} highlights our proposed approach. 

Our main contributions are as follows: 
\begin{itemize}[noitemsep]
    \item We introduce CoSS, a space similarity inclusive novel objective which motivates the student to mimic its teacher's embedding structure.
    \item We discuss the limitation imposed by only relying on $L_2$ normalized features for learning the manifold.
    \item The simplicity of our approach does not impede the final performance of trained students. We report state of the art results on various UKD benchmarks
\end{itemize}

The structure of the paper is as follows. In the subsequent \cref{sect:related_work}, we delve deeper into prior work. In \cref{sect:motivation}, we motivate the importance of space similarity. In \cref{sect:method}, we provide the details of our simple yet effective method.  In \cref{sect:experiments}, we report results on various benchmarks. We discuss the implications of our work and findings in \cref{sect:discussion}. Lastly, we conclude with final remarks in \cref{sect:conclusion}.

\section{Related Work}
\label{sect:related_work}
\subsection{Logit Based Distillation}
Early solutions for distillation have been designed for a fully supervised teacher and leverage the output space of the teacher model. Soft-label distillation ~\cite{kd_sld} is amongst the first work towards training smaller networks with guidance from larger teacher networks. Apart from supervised loss from the ground-truth labels, it minimises cross-entropy between the teacher's and the student's output logit distribution. Many methods since then have utilised the output logits to improve the knowledge distillation performance~\cite{zhang2018deep,yang2019snapshot,cho2019efficacy,mirzadeh2020improved,jin2023multi}.  

\citet{kdst} propose a correlation based formulation which is very similar to ours. The key difference apart from the teacher (supervised vs. unsupervised) is that they normalise the logits (via. softmax) and then compute inter and intra-class similarities. Whereas, we independently normalize spatial and feature dimensions of the embedding features. From the perspective of computing intra-class similarity, it is logical to apply the softmax beforehand for generating class-wise scores, however, when operating on the embedding space, any normalisation on the features alters the space information as well. 

\subsection{Feature Based Distillation}
Feature based distillation approaches leverage the internal representations produced by the teachers and the students.

Many novel approaches have incorporated internal layers, attention mechanisms etc.\ to match various novel objectives between student and the teacher~\cite{kd_fitnets, kd_at, kd_fsp, kd_mmd, kd_ft, kd_lit, kd_sp, kd_vid, hao2023oneforall}. However, as most of these approaches require careful selection of an appropriate statistic, it can be a drawback in practice for defining the distillation procedure for newer architectures. Many existing approaches also utilise local and global relationships~\cite{kd_lmt, kd_cc, kd_rkd, kd_pkt, kd_darkrank} in a metric learning framework. They mostly differ in the devised relationships for optimisation. LP \cite{kd_lp} further develops the idea of FitNets and introduces locality-preserving loss, which relies on identifying K-nearest neighbours within the training batch. Lastly, there are also methods which utilize self-supervision for knowledge distillation. CRD~\cite{kd_crd}, WKD~\cite{kd_wkd} and SSKD \cite{kd_sskd} fall into this category. A common theme for all these methods is that they focus on distillation of a teacher trained with supervision and thus often operate in conjunction with a supervised objective. Though they can be utilised for performing distillation of an unsupervised teacher, they have not yet been evaluated in such a setting. A quick evaluation (see supplementary document) of these methods shows that they do not readily adapt to a fully unsupervised setting.

SEED~\cite{kd_seed} is the first work, to the best of our knowledge, that attempts to distill knowledge in an unsupervised teacher. They perform knowledge distillation of a self-supervised teacher by minimizing the divergence between the similarity response of teacher and student on a common embedding queue. CompRess~\cite{kd_compress} introduces two feature queues, one each for the teacher and the student. SimReg~\cite{kd_simreg} demonstrates the applicability of a pairwise mean squared error minimisation between the latent embeddings of teacher and student models. Our work leverages and extends the pairwise feature matching objective for enforcing manifold similarity. AttnDistill~\cite{attnDistill} is a similar non-contrastive UKD method which focuses solely on vision transformers~\cite{vit}. It leverages the multi-head self-attention mechanism during training. DisCo~\cite{kd_disco} performs a consistency regularization between augmented versions of the input in addition to unsupervised distillation. 
BINGO~\cite{kd_bingo} is a two-stage method for performing unsupervised distillation. In the first stage it computes k-nearest neighbours for each training sample in order to construct a bagged dataset. In the second (distillation) stage, it adopts a contrastive distillation approach to minimise divergence between samples from the same bag while increasing distance w.r.t samples from different bags. SMD \cite{kd_smd} focuses on mining hard positive and negative pairs rather than operating on all pairs for distillation. To counter the influence of wrongly assigning positive and negative labels, it utilizes a weighting strategy.   
PCD~\cite{kd_pcd}, a recently proposed contrastive method, specializes in dense prediction tasks like image segmentation and object detection.

\subsection{Key Differences}
Our proposed method is designed for unsupervised distillation, but we believe it stands out from existing methods in both supervised and unsupervised regimes, despite its simplicity. In particular, our approach focuses on directly motivating the student to learn its teacher's latent manifold. As a quantitative summary, CoSS differs from many existing UKD methods in the \textbf{absence} of 
\begin{enumerate*}[label=(\roman*)]
\item feature queues,
\item contrastive objectives,
\item and, heavy augmentations.
\end{enumerate*}

\section{Motivation}
\label{sect:motivation}
In deep learning, it is generally accepted that neural networks learn a mapping from high dimensional space to a low dimensional manifold. Moreover, assumption of a locally euclidean manifold in unsupervised learning is fundamental but rarely articulated. 

For example, many unsupervised learning methods employ manifold learning based data visualisations\citep{ssl_cpc, unsup_collab}. These manifold learning approaches 
assume the embedding manifold to be locally eucldiean\citep{ml_survey, ml_tsne}. 
This assumption of the manifold being locally euclidean, allows us to treat the embedded manifold as a topological manifold. Here, we present a simple argument to show that methods solely relying on $L_2$ normalized features cannot learn the teacher's embedding structure reliably.

\begin{definition}
 Two topological spaces $\mathcal{X},\; \mathcal{Y}$ are homeomorphic if there exists a mapping $f: \mathcal{X} \rightarrow \mathcal{Y}$ s.t. $f$ is continuous, bijective and its inverse function $f^{-1}$ is also continuous.  
\end{definition}

Homeomorphism\citep{homeo} defines the concept of similarity (equivalence) between the two topological spaces. For methods which only rely on normalized cosine similarity, the student's normalized manifold and the original teacher's manifold are not homeomorphic. This is because the operation of $L_2$ normalisation is not a homeomorphism. It is not continuous, bijective, and lacks a continuous inverse. A straight forward example to support it is that, after normalization, all points lying on the ray starting from the origin will be mapped onto the same point on a hypersphere. Hence, minimisation of an objective operating on the normalized space will not preserve the original \textbf{un-normalized} structure. Existing UKD methods of SEED, DisCo, and BINGO leverage the normalized features during their intermediate steps while SMD, SimReg and AttnDistill perform regression on the normalized feature as an additional training objective. Since, the normalisation erases the original structure, an alternative strategy to retain the spatial information is by preserving similarity along the spatial dimensions. 

\section{Method}
\label{sect:method}
Our motivation centers around the idea of imposing \textit{homeomorphism} between the manifolds of a teacher and student. In order to achieve this, we propose a two step training method. In the first (offline) step, we compute k-nearest neighbours for the training samples. In the second stage, we perform distillation.

\subsection{Offline Pre-processing}

Typically, during training, a mini-batch is often composed of randomly selected samples without replacement from the training set $\mathcal{X} = \{x_1, x_2, \dots, x_N\}$. The local structure of the manifold cannot be well established if the local neighbourhood information is missing.

Hence, in-order to utilise local neighbourhood information during distillation, we append randomly selected $k$ samples (without replacement) for each $x_i$ in a mini-batch from the k-nearest set $\Omega^k_i$. 

Computation of $\Omega^k_i$ is an offline process and is performed prior to the distillation step. Utilizing the teacher model $f_t$, we first compute a similarity matrix $\mathbf{S} \in \mathbf{R}^{N\times N}$. Where, $S_{ij} = \hat{f}_t(x_i) \cdot \hat{f}_t(x_j) \; \forall \; \mathcal{X}$ and $\hat{.}$ denotes a $L_2$ normalized vector. The embeddings, $f_t(x) \in \mathcal{R}^{d}$, is the response gathered at the penultimate layer of a network (typically after global average pooling~\citep{nin}). We define the nearest neighbourhood set $\Omega_i^k$ as: 
\begin{equation}
    \Omega_i^k = \argmax(S_{i\cdot}, k),
\end{equation}
where $\argmax(\cdot, k)$ returns the indices of top k items in $S_{i\cdot}$. We shall be using these pre-computed nearest neighbour sets during the distillation process.

\subsection{Training Objectives}

\begin{wrapfigure}[20]{r}{0.3\textwidth}
  \begin{center}
    \includegraphics[width=1\linewidth]{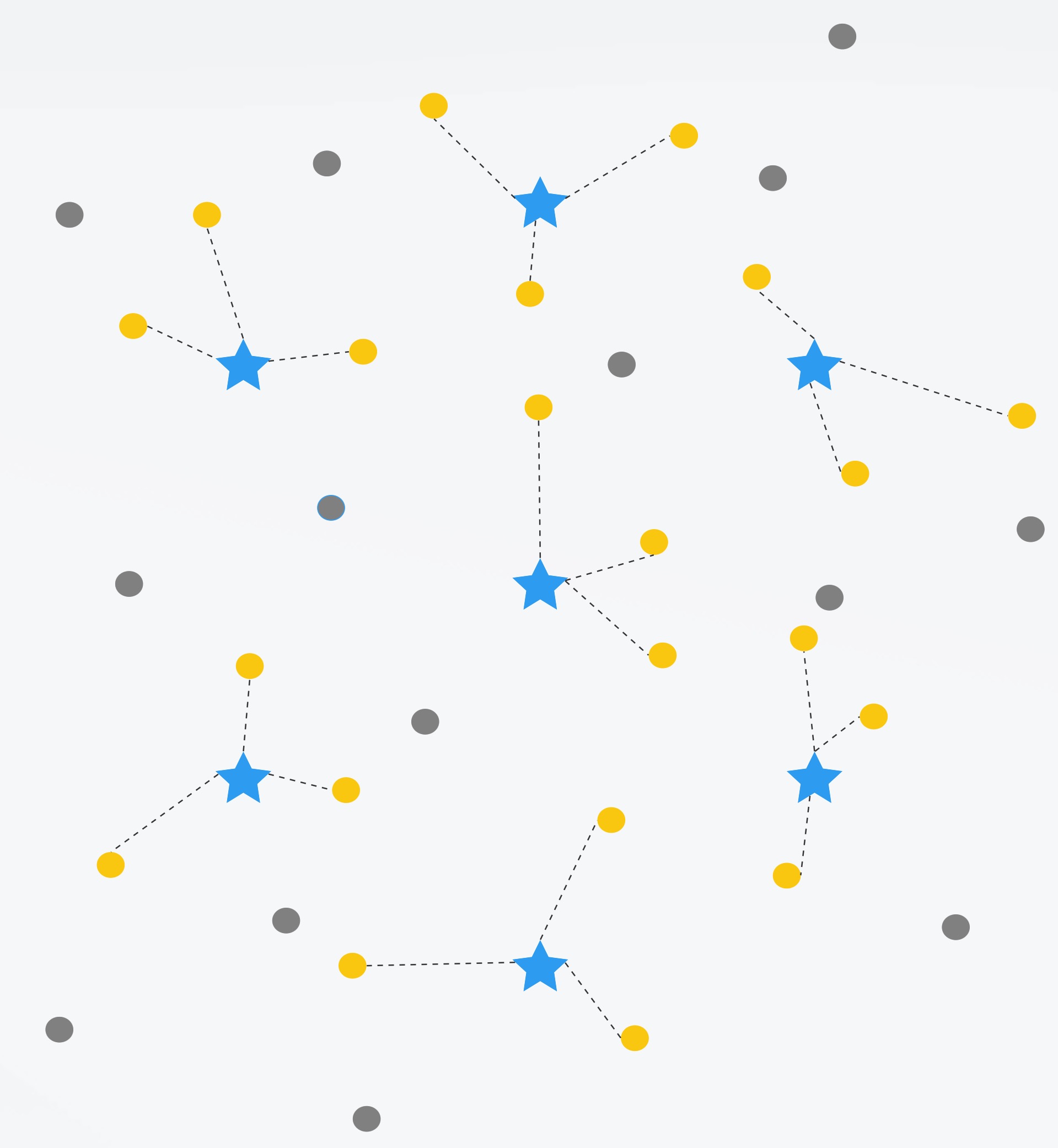}
  \end{center}
  \caption{The training batch is composed of random samples ($\mathcolor{blue}\star$) and their nearest $k$ samples ($\mathcolor{YellowOrange}\bullet$).}
  \label{fig:sampling}
\end{wrapfigure}

Similar to the teacher, we define the student neural network as $f_s$ with the output dimension as $d_s$. If $d_s \neq d_t$, we can use a small projection head for the student which can then be discarded after the distillation process \citep{kd_seed, kd_bingo, kd_disco}. We thus replace $d_t$ and $d_s$ by $d$. 

Firstly, for a mini-batch $B$, we append additional samples to it from $\Omega^k_i \; \forall \; i=1,2,\cdots |B|$ and denote the updated mini-batch as $\bar{B}$ (\Cref{fig:sampling} illustrates the enhancement of the training batch). We denote the embedding representations of all inputs in $\bar{B}$ generated by the teacher and student as matrices $A_t$ and $A_s \in R^{bk\times d}$ respectively. Here, $A^{i}_\cdot$ is the embedding output $f_\cdot(x_i)$. We then compose a matrix of normalized feature vectors $\hat{A_\cdot} = [\hat{A}_\cdot^0, \hat{A}_\cdot^1, \dots \hat{A}_\cdot^{bk}]^T$. The widely known normalized cosine similarity loss on features is defined as:
\begin{align}
 \mathcal{L}_{co} = - \frac{1}{bk}\sum_{i=0}^{bk} cosine(\hat{A_s^i},\;\hat{A_t^i}).
\end{align}
The loss computes the cosine similarity between corresponding embeddings of the teacher and student thus, performing a pairwise alignment of features in the normalized embedding manifold. Due to its simplicity and effectiveness $\mathcal{L}_{co}$ is employed by a number of existing UKD methods~\cite{kd_smd,kd_simreg,attnDistill}. 

To define space similarity, we first construct the transpose matrix of features $Z_\cdot = A_\cdot^T$ and its normalized version $\hat{Z_\cdot}$. The space similarity loss is
\begin{align}
 \mathcal{L}_{ss} = - \frac{1}{d}\sum_{i=0}^{d} cosine(\hat{Z_s^i},\;\hat{Z_t^i}).
\end{align}
Note, minimizing the loss along the spatial dimension indeed imposes \textit{homeomorphism} as the normalization here scales all data points identically. In case of minimum space alignment loss, $f_s(x_i) = \frac{\boldsymbol{\alpha}}{\boldsymbol{\beta}} f_t(x_i)$ where $\boldsymbol{\alpha}$ and $\boldsymbol{\beta}$ are scaling (normalization) vectors for the teacher's and student's dimensions respectively. The scaling components for dimensions are not required to be identical, hence, the alignment is similar for each dimension up to a scale. As $\boldsymbol{\alpha} > \mathbf{0}$ and $\boldsymbol{\beta} > \mathbf{0}$, the mapping imposed by $\mathcal{L}_{ss}$ between the corresponding projected data points is continuous, bijective and invertible.  

The loss is also very simple to implement and requires one to transpose the feature matrices prior to the normalization. Our final objective is composed of weighted combination of \textbf{Co}sine similarity and \textbf{S}pace \textbf{S}imilarity: 
\begin{equation}
    \mathcal{L}_{CoSS} = \mathcal{L}_{co} + \lambda \mathcal{L}_{ss}
\end{equation}

\cref{fig:coss} provides a pictorial representation of our approach. Feature similarity compares the features being extracted for a single sample by the teacher and student whereas, space similarity compares corresponding spatial dimensions.

\section{Experiments}
\label{sect:experiments}

\subsection{Settings}
\label{sect:settings}
\textbf{Dataset: }
Following SEED, BINGO and others, we report top-1/5 and knn-10 classification accuracy on the ImageNet~\cite{data_imagenet} dataset. For transfer learning, we utilise CIFARs\citep{dataset_cifar}, STL-10 \citep{dataset_stl10}, Caltech-101 \citep{dataset_caltech}, Oxford-IIIT Pets \citep{dataset_pets}, Flowers \citep{dataset_flowers} and DTD \citep{dataset_dtd}. For dense prediction tasks, we employ the PASCAL VOC (\texttt{trainval2007, trainval2012, test2007})~\cite{dataset_pascal} and MS-COCO (\texttt{train2017, val2017})~\cite{dataset_coco} datasets. For image retrieval experiments, we employ the Oxford-5k~\cite{oxford5k} and FORB~\cite{forb} datasets. We utilize various ImageNet variants namely, ImageNet-v2 \cite{dataset_imagenet-v2}, ImageNet-Sketch \cite{dataset_imagenet-s} and ImageNet-C \cite{dataset_imagenet-c} to understand robustness of distilled networks.

\textbf{Teachers: }
Following SEED, we use the ResNet-50 model pretrained on ImageNet using Moco-v2~\cite{mocov2}. For dense prediction, following PCD, we use the ResNet-50 pretrained on ImageNet using Moco-v3~\cite{mocov3}. 

\textbf{Students: }
We use ResNet-18, ResNet-34, and EfficientNet-B0 as the student architectures. They consist of 10.7M, 20.4M and 4M parameters respectively.

\textbf{Distillation: }
We discard the projection head of the teacher and work directly with the $2048$ dimensional features from ResNet-50 and add a projection head on top of the students to match the final embedding dimensions. Using $k\!=\!15,\ N\!=\!31, \ B\!=\!64, \ \lambda\!=\!1.0$, an initial learning rate of $0.03$ and a cosine scheduler for learning rate decay, we perform distillation for $25$ epochs distributed over $4$ GPUs. We use the augmentation policy `mocov2\_aug' as described in SEED. We also scale the overall loss by $70.0$, as we observed the convergence to be slow.  

\subsection{Supervised Classification}
\label{sect:imagenet}
\begin{table*}[t!]
\centering
\captionsetup{font=small}
\caption{\textbf{Unsupervised distillation of a (self-supervised) ResNet-50 teacher on ImageNet-1K.} The \textit{teacher} and \textit{student} correspond to models trained using Moco-v2. Values with $\diamond$ are replicated results using the corresponding official implementations. We highlight the best student performance in \textbf{bold}}
\setlength{\tabcolsep}{3pt}
    \centering
    \small{\resizebox{\textwidth}{!}{\begin{tabular}{lccccccccc}
\toprule
Teacher (top-1) &\multicolumn{9}{c}{67.40}\\ 
\midrule
\multirow{2}{*}{Methods} &\multicolumn{3}{c}{ResNet-18}&\multicolumn{3}{c}{ResNet-34}&\multicolumn{3}{c}{Eff-b0}\\
\cmidrule(lr){2-4}\cmidrule(lr){5-7}\cmidrule(lr){8-10}
 &T-1 & T-5 & KNN-10  & T-1 & T-5 &  KNN-10 & T-1 & T-5 & KNN-10\\
\midrule    
Moco-v2 & 52.20& 76.60 & 36.70&56.80&81.40 & 41.50 & 42.20 & 68.50 & 30.0\\
\midrule
SEED~\cite{kd_seed}&57.60& 81.80 & 50.12&58.50&82.60& 45.20&61.30&82.70&53.11\\
BINGO~\cite{kd_bingo} & 61.40 & 84.30 &\textbf{54.16}$^\diamond$ &63.50&85.70& -- &63.74$^\diamond$ & 85.32$^\diamond$ &54.75$^\diamond$ \\
DisCo~\cite{kd_disco} &  60.60 & 83.70 & 52.03 & 62.50 & 85.40 & 53.65 & 66.50 & 87.60 & 54.78 \\
SMD$^\diamond$~\citep{kd_smd} & 59.56 & 83.29 & 49.69 & 62.75 & 85.25 & 52.61 & --& -- & -- \\
\noalign{\smallskip}
\hdashline
\noalign{\smallskip}
 \textbf{CoSS (Ours)} & \textbf{62.35} & \textbf{84.81} & 53.78 & \textbf{64.01} & \textbf{86.14} & \textbf{54.80} & \textbf{67.36} & \textbf{87.75} & \textbf{58.33}  \\
\bottomrule
\end{tabular}}}
\label{tab:imagenet}
\end{table*}

We evaluate the goodness of a UKD method by performing classification leveraging the embedding features from the trained student network. For the ResNet baselines of BINGO, SEED, and DisCo, we present findings as reported in their corresponding publications. Since BINGO does not provide results for EfficientNet, we report the reproduced values for BINGO from the official implementation using recommended hyper-parameters. Regarding SMD, their evaluation involves a distinct teacher-student pairing; hence, we employ their official implementation and recommended hyper-parameters to conduct distillation using our selected teacher-student pair. As SimReg is distillation with $\mathcal{L}_{co}$ objective without nearest neighbour sampling, we defer the comparison to the adblation sstudies in \cref{sect:ablation}. We do not compare our work with AttnDistill as it is a solely vision transformer based approach which utilises extended training (up to 800 epochs). 

\textbf{Linear Evaluation: }Following the precedent set by SEED, we freeze the backbone and only learn the final linear classification layer. We use a batch size of $256$ and an initial learning rates of $10$ and $3$ for ResNets and EfficientNet respectively. We perform this fine-tuning for 100 epochs with learning rate reduced by a factor of $10$ at $60$ and $90$ epochs.

As per the results in \cref{tab:imagenet}, CoSS achieves significant improvements over the Moco-v2 trained (Student) baseline both for top-1 and top-5 classification accuracies. Compared to the Moco-v2 student, for top-1 accuracy, our method provides an improvement of $10\%$, $7\%$ and $15\%$ for ResNet-18, ResNet-34 and EfficientNet-b0 respectively. Comparing CoSS to state of the art methods of BINGO and DisCo, we can also observe CoSS' overall improved performance. For top-1, CoSS consistently provides gains over the closest second student. EfficienetNet-b0, while only possessing $16.3\%$ of the parameters, top-1 results are comparable to the teacher with a difference of only $0.04\%$.  

\textbf{K-NN evaluation: } As per prior studies, we report the result of \textit{k} nearest evaluation~\cite{kd_bingo, kd_seed}. Following SEED, we choose 10 nearest neighbours from the training samples and assign the majority class as the prediction. 

As shown in table \cref{tab:imagenet}, CoSS consistently achieves either state of the art or competing performance for k-NN. We believe that the performance gain of CoSS is a result of direct modelling of the teacher's manifold.

\subsection{Transfer Learning}
\label{sect:transfer}
We employ the transfer learning benchmark \cite{transfer_benchmark} to evaluate the transferability of learned representations. In this evaluation, we keep the student backbone frozen and exclusively train the final classification layer. Ensuring fairness in evaluation, we conduct a thorough hyper-parameter sweep across various learning configurations to pinpoint the optimal set of values for each model across each dataset.

As depicted in \cref{tab:transfer}, CoSS students exhibit superior transfer learning accuracies across a wide array of datasets for both ResNet-18 and EfficientNet-b0 architectures. It yields the best top-1 accuracies for $6$ and $7$ datasets for ResNet-18 and Efficient-B0 students respectively. On STL-10, it provides competitive performance to the baselines. 

\begin{table*}[t!]
\centering
\captionsetup{font=small}
\caption{\textbf{Transfer learning evaluation of distilled ResNet-18 and Efficientnet-b0.} Here, we report the top-1 accuracy.}
\setlength{\tabcolsep}{4pt}
    \centering
    \footnotesize{ \resizebox{\textwidth}{!}{\begin{tabular}{lccccccccccccccc}
\midrule
Method & \multicolumn{2}{c}{\text{CIFAR-10}}&\multicolumn{2}{c}{\text{CIFAR-100}}&\multicolumn{2}{c}{\text{STL-10}}&\multicolumn{2}{c}{\text{Caltech-101}}&\multicolumn{2}{c}{\text{Pets}}&\multicolumn{2}{c}{\text{Flowers}} & \multicolumn{2}{c}{\text{DTD}} \\
\cmidrule(lr){2-3}\cmidrule(lr){4-5}\cmidrule(lr){6-7}\cmidrule(lr){8-9}\cmidrule(lr){10-11}\cmidrule(lr){12-13}\cmidrule(lr){14-15}
& ResNet-18 & Eff-b0 & ResNet-18 & Eff-b0 & ResNet-18 & Eff-b0 & ResNet-18 & Eff-b0 & ResNet-18 & Eff-b0 & ResNet-18 & Eff-b0 & ResNet-18 & Eff-b0\\
\midrule
SEED & 85.27 &88.85& 62.75 &69.87& 93.99 &94.90& 80.26 &84.98& 76.18 &78.81& 75.10 &88.44& 67.34&69.79\\
BINGO & 87.67 &89.74& 66.14 &70.25& 94.99 &94.75& 83.84 &86.48& 79.24 &80.80& 83.62 &90.35& 70.00&70.85 \\
DisCo & 88.11 &91.63& 67.50 &73.97& \textbf{95.04} &\textbf{95.78}& 84.65 &86.44& 77.86 &81.20& 83.69 &89.52& 69.89&71.91\\
SMD & 86.47 & -- & 64.42 & -- & 94.24 & -- & 80.59 &--& 74.59 &--& 78.97 &--& 69.31 & --\\
\noalign{\smallskip}
\hdashline
\noalign{\smallskip}
CoSS (ours)& \textbf{89.23} & \textbf{92.72} & \textbf{70.11} & \textbf{77.17} & 94.11 & 95.41 & \textbf{86.62} & \textbf{90.26} & \textbf{79.98} & \textbf{83.09} & \textbf{85.78} & \textbf{93.52} & \textbf{70.43} & \textbf{74.04}\\
\bottomrule
\end{tabular}}}
    \label{tab:transfer}
\end{table*}

\subsection{Dense Predictions}
\label{sect:dense}
As previously introduced, PCD is a UKD method aimed at learning a student model focused on dense prediction tasks of object detection and image segmentation. For this evaluation, we follow the protocols as outlined by PCD. We utilize the student ResNet-18 distilled from a Moco-v3 ResNet-50 teacher~\cite{mocov3}. The hyper-parameters for distillation remain consistent with those mentioned earlier (see \cref{sect:settings}). We use the Detectron2~\cite{detectron2} package for training and evaluation. The baseline results are directly sourced from PCD.

On VOC2007+2012~\cite{dataset_pascal} we train C4 backbone with Faster RCNN~\cite{faster_rcnn} detector whereas for MS-COCO~\cite{dataset_coco}, we train C4 with a Mask R-CNN~\cite{mask_rcnn}. For VOC2007+2012, we set the initial learning rate at $0.1$ and perform the training for $24k$ iterations and evaluate the results on \texttt{test2007} set. On COCO~\cite{dataset_coco}, following PCD, we fine-tune the detector with an initial learning rate of $0.11$ for $90k$ iterations. The corresponding scheduling for both the trainings are also referred to as $1\times$ in Detectron2's documentation. For detailed hyper-parameters such as warm up, augmentations, learning rate decay and image resolutions etc. we refer the readers to PCD. 

Results in \cref{tab:dense} show that the CoSS student is able to compete with PCD across various settings even though, it is not specifically designed for dense predictions. On the more challenging dataset of COCO, CoSS outperforms PCD. We believe the strong performance of CoSS comes from the overall better knowledge distillation performance. Comparing, top-1 accuracy of the models on ImageNet, we can observe that CoSS provides $\approx 2\%$ gain over PCD.
\begin{table*}[t]
\centering
\captionsetup{font=small}
\caption{\textbf{Object Detection and Segmentation evaluation of ResNet-18 students.} Values for baselines are reported directly from PCD.}
\setlength{\tabcolsep}{4pt}
\footnotesize{\begin{tabular}{lccccc}
\toprule
Method & ImageNet & \multicolumn{2}{c}{VOC 07+12}  & \multicolumn{2}{c}{MS-COCO} \\
\cmidrule(lr){3-4}\cmidrule(lr){5-6}
& T-1  & AP$_\text{50}$ & AP & AP$^\text{bbox}$ & AP$^\text{mask}$ \\
\midrule
Teacher & 74.6 &  83.0 & 56.7 & 37.4 & 32.8 \\
\midrule
CompRess~\cite{kd_compress} & 63.9 &  78.4 & 50.4 & 31.4 & 28.4 \\
DisCo & 63.5 &  72.6 & 40.1 & 28.2 & 25.8  \\
BINGO & 64.2 &  77.8 & 49.3 & 31.1 & 28.2 \\
PCD~\cite{kd_pcd} & 65.1 & \textbf{79.4} & 52.1 & 32.2 & 29.0 \\
\noalign{\smallskip}
\hdashline
\noalign{\smallskip}
 \textbf{CoSS (Ours)} & \textbf{67.2} & 78.7 & \textbf{52.4} & \textbf{33.9} & \textbf{30.2}\\
\bottomrule
\end{tabular}
}
\label{tab:dense}
\end{table*}

\subsection{Image Retrieval}
\label{sect:image_ret}
In this experiment, we investigate the efficacy of knowledge distillation models in the context of image retrieval. Oxford-5k is a widely used dataset for image retrieval. It consists of $55$ query images and $\approx 5000$ database images categorized into \textit{medium} and \textit{hard} based on the perceived difficulty of the task. FORB is a recently proposed dataset for image retrieval in the flat object setting. It consists of $8$ categories with a total of $14,000$ query images and $54,000$ database images. The queries are labelled as \textit{easy}, \textit{medium} and \textit{hard} based on the difficulty level. For Oxford-5k, we report the mean average precision for \textit{medium} and \textit{hard} instances. Whereas, for FORB, we report the overall mean average precision across \textit{easy}, \textit{medium} and \textit{hard} categories.
We use the ResNet-18 student distilled from Moco-v2 ResNet-50 for this evaluation.

\Cref{tab:retrieval} presents the results of the retrieval task. According to these results, CoSS demonstrates improved mAP performance compared to the baseline models. For the few instances where CoSS is not the top performing model, it achieves a second best rating. These results suggest the potential effectiveness of the latent representations produced by the student across various scenarios, indicating promising capabilities for real-world applications.

\begin{table*}[t!]
\centering
\captionsetup{font=small}
\caption{\textbf{Image retrieval performance on Oxford-5k and FORB, measured in mAP.} We highlight the best student performance in \textbf{bold}}
\setlength{\tabcolsep}{4pt}
    \centering
    \small{\footnotesize{ \resizebox{\textwidth}{!}{\begin{tabular}{lcccccccccc}
\toprule
\multirow{2}{*}{Methods} &\multicolumn{2}{c}{Oxford-5k}&\multicolumn{8}{c}{FORB}\\
\cmidrule(lr){2-3}\cmidrule(lr){4-11}
 & Medium & Hard  & Animated & Photorealistic & Bookcovers & Paintings & Currency & Logos & Packaged & Posters \\
\midrule
SEED  & 17.19 & 3.71 & 0.040 & 0.184 & 0.099 & 0.267 & 0.108 & 0.007 & 0.071 & 0.100\\
SMD   & 16.77 & 3.53 & 0.043 & 0.133 & 0.093 & 0.214 & 0.121 & 0.006 & 0.065 & 0.102\\ 
DisCo & 18.63 & 4.12 & 0.120 & 0.283 & 0.176 & 0.336 & 0.215 & \textbf{0.016} & 0.129 & 0.160\\
BINGO & 19.80 & \textbf{5.84} & 0.158 & 0.299 & 0.193 & 0.305 & 0.243 & 0.012 & 0.144 & 0.172\\
\noalign{\smallskip}
\hdashline
\noalign{\smallskip}
CoSS & \textbf{20.59} & 5.50 & \textbf{0.179} & \textbf{0.302} & \textbf{0.220}  & \textbf{0.353} & \textbf{0.286} & 0.013 & \textbf{0.155} & \textbf{0.205} \\
\bottomrule
\end{tabular}}}}
\label{tab:retrieval}
\end{table*}

\subsection{Out-of-Distribution Robustness}
\label{sect:ood}
\begin{table*}[t!]
\captionsetup{font=small}
 \caption{\textbf{Robustness evaluation of Moco-v2 ResNet-50 distilled ResNet18s.} We report the Top-1 classification accuracy. We highlight the best performance in \textbf{bold}.}
    \centering
    \resizebox{\textwidth}{!}{%
    \begin{tabular}{lcccccccccccccccccccc}
\midrule
\multirow{2}{*}{Method}&\multicolumn{3}{c}{\text{ImageNet-v2}}&\multirow{2}{*}{\text{ImageNet-S}}&\multicolumn{15}{c}{\text{ImageNet-C}}\\
&MF&Tr&Top&&brightness & contrast & defocus & elastic &  fog & frost  & gaussian & glass &  impulse &  jpeg &  motion & pixelate & shot &  snow  & zoom\\
\midrule
SEED & 45.46 & 53.24 & 60.25 & 13.36 & 53.84 & 48.33 & 36.75 & 48.64 & 36.65 & 39.74 & 39.85 & 33.86 & 23.86 & 41.38 & 39.32 & 47.14 & 37.81 & 30.97 & 26.67\\
BINGO & 46.80 & 55.58 & 62.21 &14.49 &55.51 & 49.91 & 37.45 & 49.97 & 37.96 & 41.51&40.17 & 33.85 & 23.70 & 44.10 & 39.18 & 49.39 & 38.77 & 32.20 & 27.33\\
DisCo & 47.66 & 55.96 & 62.74 & 14.40 & 56.07 & 50.79 & 39.05 & 50.56 & 38.18 & 42.32 & 42.39 & 36.34 & 27.24 & 44.83 & 41.84 & 50.10 & 41.12 & 33.27 & 28.43\\
SMD & 47.13 & 55.40 & 61.84 & 13.15 & 55.51 & 50.18 & 38.72 & 50.50 & 38.12 & 41.12 & 40.44 & 34.66 & 24.60 & 44.22 & 40.66 & 49.10 & 38.92 & 32.39 & 28.60\\
\noalign{\smallskip}
\hdashline
\noalign{\smallskip}
CoSS (ours)& \textbf{49.70} & \textbf{58.21} & \textbf{64.84} & \textbf{15.90} & \textbf{58.78} & \textbf{53.26} & \textbf{42.44} & \textbf{53.49} & \textbf{42.09} & \textbf{44.72} & \textbf{45.32} & \textbf{38.74} & \textbf{27.74} & \textbf{48.82} & \textbf{44.42} & \textbf{53.42} & \textbf{43.54} & \textbf{35.58} & \textbf{31.79}\\
\bottomrule
\end{tabular}}
    \label{tab:ood}
\end{table*}
Various studies have indicated that deep learning systems often break down when encountered with data outside the training distribution\citep{dataset_imagenet-c, dataset_style, dataset_imagenet-s}. Due to the wide spread applicability of deep learning systems in the real world, it becomes important to ensure a high degree of robustness of such systems. In this experiment we explore the robustness of trained students under different kinds of shifts in the input data. ImageNet-v2 is a natural image dataset which closely resembles the sampling process of the original ImageNet dataset. It has been employed by various studies to understand the robustness of models under natural data shift~\citep{natural_shift, ood_shift_benchmark}. ImageNet-S consists of sketch images for ImageNet classes. It is a stronger deviation from the types of images present in the training set. ImageNet-C is a synthetically generated dataset for evaluating a model's performance under various forms of corruptions. We utilise corruption=1 for our evaluation.

Results reported in \cref{tab:ood} demonstrate that CoSS is robust across various input distribution shifts. It even outperforms BINGO which employs stronger data augmentations in the form of CutMix~\cite{cutmix}. Strong data augmentations such as CutMix have been shown to improve model robustness. 

\subsection{Qualitative Evaluation}
\label{sect:toy}

\begin{figure}[t!]
  \centering
  \begin{subfigure}{0.16\textwidth}
    \includegraphics[width=\textwidth]{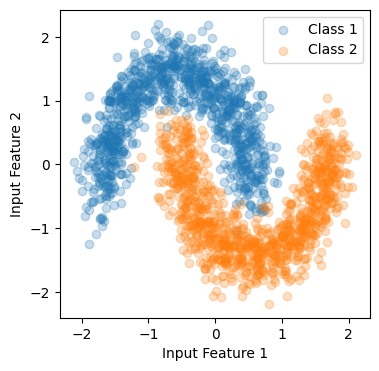}
  \end{subfigure}
  \begin{subfigure}{0.16\textwidth}
    \includegraphics[width=\textwidth]{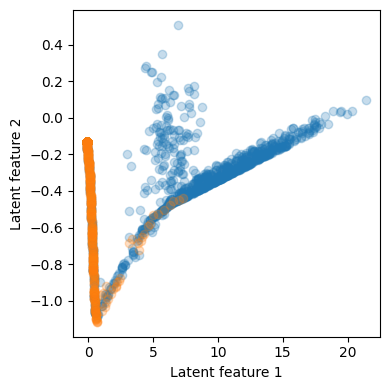}
  \end{subfigure}
  \begin{subfigure}{0.16\textwidth}
    \includegraphics[width=\textwidth]{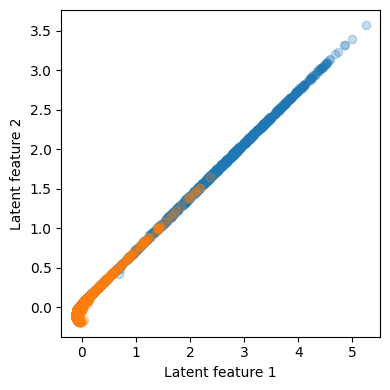}
  \end{subfigure}
  \begin{subfigure}{0.16\textwidth}
    \includegraphics[width=\textwidth]{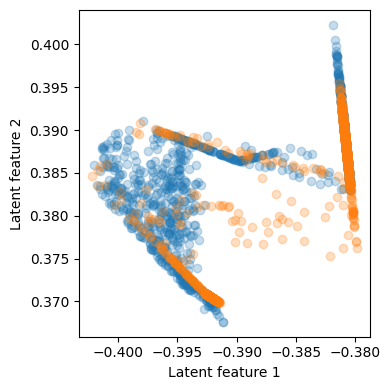}
  \end{subfigure}
  \begin{subfigure}{0.16\textwidth}
    \includegraphics[width=\textwidth]{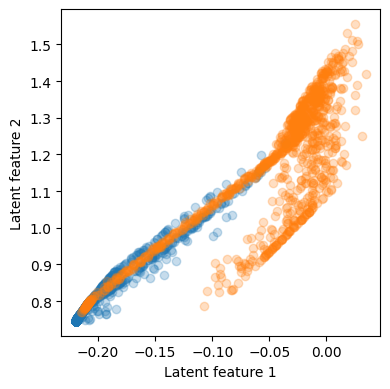}
  \end{subfigure}
    \begin{subfigure}{0.16\textwidth}
    \includegraphics[width=\textwidth]{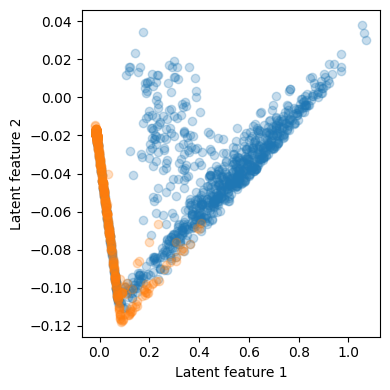}
  \end{subfigure}
  
  \hfill
  
  \begin{subfigure}{0.16\textwidth}
    \includegraphics[width=\textwidth]{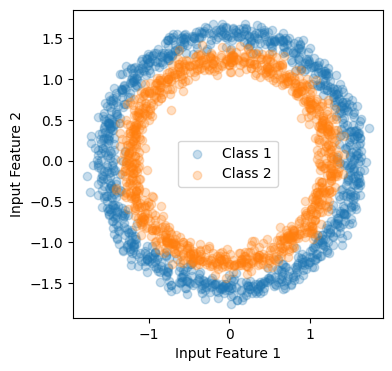}
    \caption{Data}
  \end{subfigure}
  \begin{subfigure}{0.16\textwidth}
    \includegraphics[width=\textwidth]{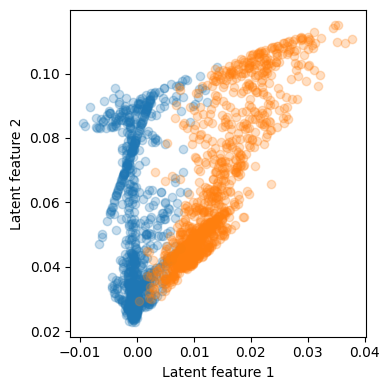}
    \caption{Teacher}
  \end{subfigure}
  \begin{subfigure}{0.16\textwidth}
    \includegraphics[width=\textwidth]{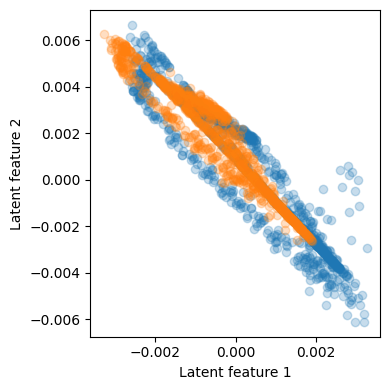}
    \caption{SEED}
  \end{subfigure}
  \begin{subfigure}{0.16\textwidth}
    \includegraphics[width=\textwidth]{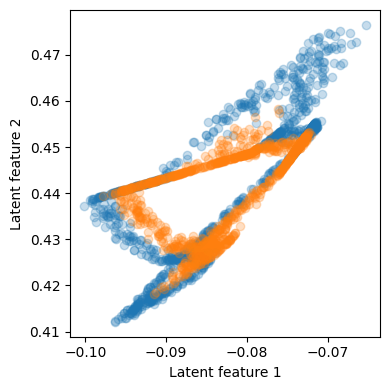}
    \caption{DisCo}
  \end{subfigure}
  \begin{subfigure}{0.16\textwidth}
    \includegraphics[width=\textwidth]{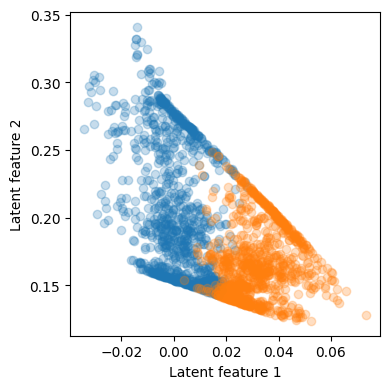}
    \caption{$\mathcal{L}_{co}$}
  \end{subfigure}
    \begin{subfigure}{0.16\textwidth}
    \includegraphics[width=\textwidth]{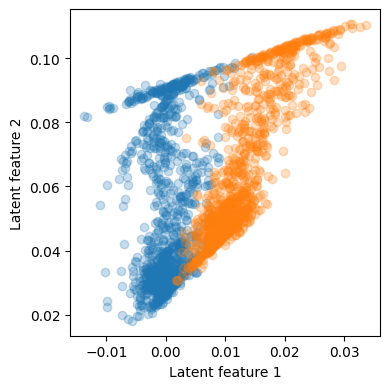}
    \caption{CoSS}
  \end{subfigure}
  \caption{\small{\textbf{Plots comparing the latent space of the teacher and different students.} Visually, we can assess that though SEED is able to separate the input samples adequately, the learnt mapping is not faithful to the teacher. Whereas, adding the space similarity objective to the standard cosine similarity allows the student to learn a mapping which aligns better with its teacher.}}
  \label{fig:toy}
\end{figure}

In this experiment, we qualitatively demonstrate the lack of manifold modelling capability of SEED, DisCo and SimReg($\mathcal{L}_{co}$) with a set of toy experiments. We first train teacher models on toy datasets (Two-moons and Circles~\cite{sklearn_api}) using a contrastive objective with positives and negatives determined by the class labels. For the architecture, we consider an MLP with 4 hidden layers, $f: R^2 \rightarrow R^2$ with ReLU activations. Hidden layers perform the following mappings $R^2\rightarrow \! R^4 \! \rightarrow \! R^8 \! \rightarrow \! R^4 \! \rightarrow \! R^2$. For the training data, we generate 2500 samples with a noise factor of $0.125$. The teacher is trained using constrastive loss for $300$ epochs with an initial learning rate of $0.001$ using an ADAM optimiser and cosine learning-rate decay. The students are trained using the same hyper-parameters. Using objectives of different students, we perform the unsupervised knowledge distillation. For DisCo, we simulate transformations to generate multiple views of the input by adding perturbations randomly sampled from $\mathcal{N}(0, 0.01)$.

In \cref{fig:toy}, we visualize the training data along with the latent spaces of teacher and student models. In the teacher's latent space we can observe few stand out structures. It can also be observed that SEED fails to adequately replicate teacher's manifold. The latent space of SEED and DisCo lacks various details in comparison to the teacher. Pairwise cosine similarity, similar to SEED and DisCo, appears to be ignoring the structure of the teacher's manifold. However, the best quantitative results are observed with cosine similarity in conjunction with space similarity. As mentioned earlier, the embedding manifold is similar up to a scale.

\subsection{Ablation Study}
\label{sect:ablation}

\begin{table}[t] 
    \centering
    \captionsetup{labelfont=bf} 
    \begin{minipage}{0.3\textwidth}
        \centering
        \caption{Examining individual loss components}
        \resizebox{1\textwidth}{!}{%
            \begin{tabular}{cccc}
        \toprule
            Arch. & $\mathcal{L}_{co}$ & $\mathcal{L}_{ss}$ & $\mathcal{L}_{coss}$ \\
            \midrule
        \midrule
         R-50/R-18 & 61.85 & 61.79 & \textbf{62.35}\\
        R50/Eff-b0 & 66.41& 66.76 &\textbf{67.36}\\
        \bottomrule
        \end{tabular}
        }
        \label{tab:ablation_components}
    \end{minipage}%
    \hspace{0.01\textwidth} 
    \begin{minipage}{0.3\textwidth}
        \centering
        \caption{Importance of nearest sampling}
        \resizebox{0.55\textwidth}{!}{%
            \begin{tabular}{ccc}
            \toprule
            $\mathcal{L}_{CoSS}^{k=0}$  & $\mathcal{L}_{CoSS}^{k=15}$ \\
            \midrule
            61.89 & \textbf{62.35} \\
            \bottomrule
            \end{tabular}%
        }
        \label{tab:ablation_k}
    \end{minipage}%
    \hspace{0.01\textwidth} 
    \begin{minipage}{0.3\textwidth}
        \centering
        \caption{Importance of $\lambda$}
        \resizebox{0.7\textwidth}{!}{%
            \begin{tabular}{ccc}
            \toprule
            $\lambda=0$ & $\lambda=0.5$ & $\lambda=1.0$ \\
            \midrule
            61.85 & \textbf{62.41} & 62.35\\
            \bottomrule
            \end{tabular}%
        }
        \label{tab:ablation_lambda}
    \end{minipage}%
    \\ 
    \begin{minipage}{0.3\textwidth}
        \centering
        \caption{Comparison to soft-label distillation}
        \resizebox{\textwidth}{!}{%
            \begin{tabular}{ccccc}
            \toprule
             & \multicolumn{2}{c}{ResNet-18} & \multicolumn{2}{c}{Eff-b0} \\
             \cmidrule(lr){2-3}\cmidrule(lr){4-5}
             & Top-1& KNN-10 & Top-1 & KNN-10 \\ 
            \midrule
            
            SLD & 59.88 & 52.67 & 61.49 & 56.10 \\
            CoSS &  \textbf{62.35} & \textbf{53.78} & \textbf{67.36} & \textbf{58.33}\\
            \bottomrule
            \end{tabular}%
        }
        \label{tab:ablation_sld}
    \end{minipage}%
    \hspace{0.01\textwidth} 
    \begin{minipage}{0.3\textwidth}
        \centering
        \caption{ResNet-101 teacher distillation}
        \resizebox{0.7\textwidth}{!}{%
            \begin{tabular}{ccc}
            \toprule
            SEED & DisCo & CoSS \\
            \midrule
            58.90 & 62.30 & \textbf{63.74} \\
            \bottomrule
            \end{tabular}%
        }
        \label{tab:ablation_teacher}
    \end{minipage}%
\end{table}
\textbf{Importance of Loss Components: }
We compare the results of distillation with individual components with the combined objective of CoSS. We report results with distillation of a Moco-v2 ResNet-50 to a ResNet-18 and EfficientNet-b0 student. The hyper-parameters are those provided in \Cref{sect:imagenet}.
Results in \cref{tab:ablation_components} highlight the effectiveness of combing both the components over a single loss objective. When using the entire dataset, the degree of improvement differs for different teacher-student architectures. In the supplementary document we share results on CIFAR-100 dataset.

\textbf{Importance of sampling: }We evaluate the impact of nearest neighbour sampling on the training of the student. Utilising ResNet-50/ResNet-18 teacher student pair we perform distillation as per the settings detailed in \Cref{sect:settings}. To recall, $k$ controls the number of nearest neighbours added to a batch of randomly sampled training images. In this experiment, we compare $k=0$ and $k=15$. $k=0$ implies that the model will be focusing more on the high level structural similarity of the manifold and may miss the finer details. We utilise 100\% of the training set for distillation. As results in \cref{tab:ablation_k} indicate, using neighbour hood sampling gives a boost in student's performance. This result highlights the importance of capturing local information when modeling the teacher's manifold in addition to its global structure.

\textbf{Value of $\lambda$: } $\lambda$ controls the contribution of space similarity for the distillation training. We have used $\lambda=1$ for our main experiments. Here, we report results with $\lambda = \{0, 0.5, 1.0\}$. $\lambda=0$ is equivalent to only applying $\mathcal{L}_{co}$ or SimReg~\cite{kd_simreg}. We use ResNet-50/ResNet-18 teacher student pair distilled as per settings outlined in \Cref{sect:settings}.
\Cref{tab:ablation_lambda} indicates that weighing the space component equally to conventional feature similarity gives improved results. Also, the performance remains consistent for $\lambda=\{0.5,1.0\}$. In the supplementary, we share results with CIFAR-100. 

\textbf{Soft-label Distillation: }In this experiment, we compare CoSS with soft-label distillation (SLD) adapted for unsupervised knowledge distillation. To recall, SLD~\cite{kd_smoothing} minimises the KL divergence between a student and teacher's output probabilities in a supervised setting. DINO~\cite{kd_dino} extends this objective for self-supervised learning. Here, we employ the same objective as that of DINO and compare it against CoSS. For SLD, the training hyper-parameters are identical to those applied for SEED and for CoSS the hyper-parameters are provided in \ref{sect:settings}. We use a ResNet-50 teacher and a ResNet-18 student for this experiment. As shown in \cref{tab:ablation_sld}, our formulation provides significant performance boost over the SLD baseline. 

\textbf{ResNet-101 teacher: }Using a ResNet-101 trained using Moco-v2 as the teacher, we perform the distillation onto a student ResNet-18. We follow the distillation hyper-parameters as outlined in \cref{sect:experiments}. For baselines, we report the values as sourced from DisCo. The findings in the \cref{tab:ablation_teacher} demonstrate that our approach readily extends to a different teacher architecture. 

\section{Discussion}
\label{sect:discussion}
In this work, we address the challenging and critical problem of unsupervised knowledge distillation. While prior studies have predominantly focused on establishing distillation-worthy relationships among samples, our approach takes a distinct perspective by directly modelling the teacher's manifold. By doing so, we aim to indirectly preserve relationships between the samples. 

Central to our approach is the utilization of space similarity to establish a form of \textit{homeomorphism} between the projections of the student and teacher models. This homeomorphism ensures that the fundamental topological properties of the learned representations are preserved during the distillation process which otherwise are lost due to sole reliance on $L_2$ normalized embedding features. Homeomorphism, however, only ensures that the alignment of individual spatial dimensions is upto a scale. Moving forward, we aim to explore even stronger constraints on learning student topologies, which could potentially improve the effectiveness and robustness of our approach. 

Our experimental results demonstrate the effectiveness of CoSS in training student models that not only perform well on the training distribution but also deliver competitive performance on various downstream tasks. For instance, CoSS outperforms PCD on the COCO dataset for dense prediction tasks. Furthermore, we believe that integrating CoSS into existing frameworks, such as PCD, could further enhance the outcomes of dense prediction tasks, opening up new avenues for research and application. 

In this study, our primary focus has been on the domain of computer vision. However, the rapid development of unsupervised large models in natural language processing presents an intriguing opportunity~\cite{future_clip, future_nlp_0, future_nlp_1, gu2024minillm}. Exploring the potential transference of our proposed method to this domain is a direction we leave for future exploration.

\section{Conclusion}
\label{sect:conclusion}
In this paper, we are inspired by the necessity to distill large models trained with self-supervised learning into smaller models. However, since the labels used to train these large models are typically unavailable, we investigate knowledge distillation in a purely unsupervised setting. In this setting, we demonstrate that unsupervised feature distillation can be achieved without the need to store a feature queue, and directly modelling the teacher's manifold.

%
%
\bibliographystyle{plainnat}
\bibliography{papers}
\newpage
\appendix
\begin{algorithm}[th]
   \caption{PyTorch based implementation}
   \label{algo:FUSS}
    \definecolor{codeblue}{rgb}{0.25,0.5,0.5}
    \lstset{
      basicstyle=\fontsize{7.2pt}{7.2pt}\ttfamily\bfseries,
      commentstyle=\fontsize{7.2pt}{7.2pt}\color{codeblue},
      keywordstyle=\fontsize{7.2pt}{7.2pt},
    }
\begin{lstlisting}[language=python,mathescape=true]
# $\color{codeblue}f_s$, $\color{codeblue}f_t$: student and teacher networks
# $\color{codeblue}\beta$: scale for the loss
# $\color{codeblue}\Omega$: nearest neighbour set
# $\color{codeblue}k$: neighbours to sample 

for x in loader: # load a minibatch x with b samples
    x = [x, $\Omega^k$]
    x = augment(x) # random augmentation
    
    $A_s$ = $f_s$(x) # student output bXd
    with torch.nograd():
        $A_t$ = $f_t$(x) # teacher output bXd
    
    $l_{co}$ = dot(A_s, A_t) #feature term
    $l_{ss}$ = dot(A_s.T, A_t.T) #space term
    loss = $\beta$*($l_{co}$ + $l_{ss}$) #feature and space terms  
    
    loss.backward() # back-propagate
    update($f_s$) # SGD
    
def dot(s, t):
    s = torch.norm(s, dim=-1)
    t = torch.norm(t, dim=-1)
    return - torch.mean((s*t).sum(dim=-1))
\end{lstlisting}
\end{algorithm}

\section{Adapting Supervised Methods}

As many distillation approaches operate on the features of the students and teachers, they can be applied to an unsupervised setting. However, to the best of our knowledge, such a study has not yet been performed. In this experiment, we make slight adjustments to the baselines implemented by CRD~\cite{kd_crd}. We remove the final-classification layer of the teacher and the student to simulate an unsupervised setting. Moreover, we set the contributions due to ground-truth and soft-labels to $0$. For a fair comparison, for the baselines, we report the best top-1 achieved from hyper-parameter sweep by scaling loss $\{0.5, 1, 2, 4, 8, 16, 32\}\times$. 

In \cref{tab:super}, we report the results of baselines and CoSS. We observe that baseline methods don't readily work in an unsupervised setting. In the future, it will be interesting to delve deeper into understanding why such a wide gap in performance exists for distillation methods when applied to an unsupervised setting.

\begin{table}[h!]
    \centering
     \caption{\small{\textbf{Unsupervised distillation performance of adapted supervised distillation methods on CIFAR-100.}}}
    \resizebox{\columnwidth}{!}{\begin{tabular}{cccccccccccc}
    \toprule
         Arch & ATT~\cite{kd_at} & SP~\cite{kd_sp} & VID~\cite{kd_vid} & RKD~\cite{kd_rkd} & PKT~\cite{kd_pkt} & Factor~\cite{kd_ft} & NST~\cite{kd_mmd} & CRD~\cite{kd_crd} & SRRL~\cite{kd_srrl} & DIST~\cite{kdst} & CoSS  \\
         \midrule
         Res20/Res56 & 49.00 & 63.14 & 68.61 & 56.28 & 61.58 & 46.63 & 25.62&65.03&69.31&67.13&\textbf{71.11} \\
         \midrule
         Vgg8/Vgg13 & 56.22 & 71.62 & 73.65 & 44.50 & 71.81 & 39.71 & 42.78 & 69.73& 72.85&73.40&\textbf{74.58} \\
         \midrule
         Res8x4/Res32x4 & 43.11 & 62.51 & 69.60 & 32.33 & 64.53 & 36.64 & 41.27 &  65.78 & 69.19& 67.67& \textbf{73.90} \\
        \bottomrule
    \end{tabular}}
    \label{tab:super}
\end{table}

\begin{table}[t]
    \centering
     \caption{\small{\textbf{Importance of different loss components.}}}
\footnotesize{
    \begin{tabular}{lccc}
\toprule
Models&$L_{co}$&$L_{ss}$&$L_{coss}$\\
\midrule
Resnet8x4/32x4 & 72.05 & 73.53 & \textbf{73.90}\\
ResNet20/56 & 70.58 & 70.42 & \textbf{71.11}\\
\bottomrule
\end{tabular}}
 \label{tab:component}
\end{table}

\section{Ablation on CIFAR-100}
\subsection{Importance of loss components}
 Following \citet{kd_crd}, we utilise the teacher models which were trained using supervision. We remove the final classification layer of the teacher to simulate an unsupervised setting. In \cref{tab:component}, we report the results of distillation on CIFAR-100 dataset. We can observe that combining feature and space similarity yields the best performing student.

\subsection{Contribution of $\lambda$}
\begin{table}[th!]
    \centering
     \caption{\small{\textbf{Contribution of $\lambda$ to distillation.}}}
    \footnotesize
     \footnotesize{\begin{tabular}{lcccc}
\toprule
 Student/Teacher&$\lambda=0$&$\lambda=0.25$&$\lambda=0.5$&$\lambda=1$\\
\midrule
 ResNet8x4/32x4 & 72.05 & 73.44 & 73.90 & \textbf{73.91}\\
ResNet20/56 & 70.58 & \textbf{71. 34}& 71.11 & 70.68\\
\bottomrule
\end{tabular}}
    \label{tab:lambda}
\end{table}

For our main experiments, we used $\lambda=1.0$. Here, we evaluate the distillation performance with different weights assigned to the space similarity component. From the reported results in \cref{tab:lambda}, we note that while combining space similarity and feature similarity yields the best-performing student, the optimal results are achieved with varying degrees of space similarity contribution. Different architectures generally demonstrate varied behaviors in response to distillation. It is not yet known why some architectures perform better than others, even when they have similar capacities. We believe that space similarity similarly impacts different architectures to a varying degree.

\section{SEED + Space Similarity}
We train the SEED objective with our proposed space similarity objective ($k=0$). We observe that the SEED + SS student achieves a top-1 of $58.30\%$ compared to the baseline SEED model's $57.60$. As we did not employ the nearest neighbour sampling, we expect the performance to further increase upon doing so. 

\section{Batch Normalisation For Space Similarity}
\label{sect:bn}
Batch Normalisation (BN) was proposed to reduce the covariate shift which occurs during the mapping of inputs from one layer to another~\citep{bn}. Since its introduction, BN has found place in numerous deep learning architectures\citep{resnetv1, resnetv2, resnext}. Here, we show how one can directly aim the distillation process to match student and network's embeddings and subsequently compare its performance with our formulation. 

BN operates on a batch of input data, $X \in R^{b\times d}$ where batch size is $b$ and $d$ is the feature dimension. It first performs standardisation $\hat{X}_{:,i} = \frac{X_{:,i} - \mu_i}{\sigma_i}$
where, $:$ denotes all the entries in the batch dimension and $\mu_i$, $\sigma_i$ are the mean and variances respectively for the $i^{th}$ feature dimension. The normalized values are then scaled by trainable parameters $\gamma_i$ and $\beta_i$ as:

\begin{equation}
    Z_{:,i} = \gamma \hat{X}_{:,i} + \beta
\end{equation}

here, $\gamma_i$ and $\beta_i$ can be interpreted as affine transformations which operate independently for different spatial dimensions. We can utilise it to map the standardised student embeddings to the teacher's unnormalized embedding space. The corresponding loss can be defined as follows: 
\[\begin{aligned}
    & \mathcal{L}_{coss} = \frac{1}{b}\sum_{i=0}^{i<b}\mathcal{D}(Z^s_i, X^t_i) \\
\end{aligned}
\]
where, $X^t_i$ is the teacher's embedding for the $i^{th}$ sample and $Z^s_i$ corresponds to the BatchNormalized student's embeddings. In table \ref{tab:bn}, we report the results using this approach on CIFAR-100 distillation task. We utilised mean-squared-error for the metric $D$.

\begin{table}[h!]
\centering
\captionsetup{font=small}
 \caption{\small{\textbf{CIFAR-100 unsupervised distillation with BN.}}}
\setlength{\tabcolsep}{5pt}
    \centering
    \footnotesize{\begin{tabular}{lccc}
\toprule
Methods&VGG13&Resnet32x/8x&WRN-40/16\\
\midrule
BN & 74.01 & 72.22  &  73.42\\
CoSS($k=0$) & 74.58  & 73.90  & 74.65\\
\bottomrule
\end{tabular}}
\label{tab:bn}
\end{table}
\end{document}